\documentclass[]{joyenfra}

\usepackage[toc,page,header]{appendix}

\usepackage{minitoc}
\usepackage{cleveref} 
\usepackage{subcaption}
\usepackage{booktabs}
\usepackage{graphicx}
\usepackage{subcaption}

\newcolumntype{L}[1]{>{\raggedright\arraybackslash}p{#1}}
\newcolumntype{C}[1]{>{\centering\arraybackslash}p{#1}}
\newcolumntype{Y}{>{\raggedright\arraybackslash}X}

\usepackage{fancyhdr} 

\usepackage{booktabs}
\usepackage{longtable}
\usepackage{tabularx}
\usepackage{array}
\usepackage{hyperref}
\usepackage{xcolor}
\usepackage{titlesec}
\usepackage{microtype}

\hypersetup{
  colorlinks=true,
  linkcolor=blue!55!black,
  urlcolor=blue!55!black,
  citecolor=blue!55!black
}

\renewcommand{\thesection}{\arabic{section}}
\renewcommand{\thesubsection}{\thesection.\arabic{subsection}}
\renewcommand{\thesubsubsection}{\thesubsection.\arabic{subsubsection}}
\titleformat{\section}{\Large\bfseries}{\thesection}{0.8em}{}
\titleformat{\subsection}{\large\bfseries}{\thesubsection}{0.8em}{}
\titleformat{\subsubsection}{\normalsize\bfseries}{\thesubsubsection}{0.8em}{}

\usepackage{natbib}
\usepackage{latexsym}

\usepackage{url}
\usepackage{amssymb}
\usepackage[utf8]{inputenc}
\usepackage{microtype}
\usepackage{booktabs}
\usepackage{pifont} 
\usepackage{multirow}
\usepackage{makecell}
\usepackage{paralist}
\usepackage{xspace}
\usepackage{color}
\usepackage{xcolor}
\usepackage{colortbl}
\usepackage{adjustbox}
\usepackage{hyperref} 
\usepackage[edges]{forest}
\usepackage{tikz} 
\usepackage{caption}
\usepackage{amsfonts}

\hypersetup{
    colorlinks,
    linkcolor={blue!80!black},
    citecolor={blue!80!black},
}
\tikzset{
    root/.style =             {align=center, text width=1cm, rounded corners=3pt, line width=0.3mm, fill=gray!10, draw=gray!80, font=\small},
    demographic/.style =         {align=center, text width=1.8cm, rounded corners=3pt, line width=0.3mm, fill=blue!10, draw=blue!80, font=\footnotesize},
    demographic_work/.style =    {align=center, text width=10cm, rounded corners=3pt, line width=0.3mm, fill=blue!10, draw=blue!0, font=\footnotesize},
    character/.style =         {align=center, text width=1.8cm, rounded corners=3pt, line width=0.3mm, fill=red!10, draw=red!80, font=\footnotesize},
    character_work/.style =    {align=center, text width=10cm, rounded corners=3pt, line width=0.3mm, fill=red!10, draw=red!0, font=\footnotesize},
    personalization/.style =           {align=center, text width=1.8cm, rounded corners=3pt, line width=0.3mm, fill=cyan!10, draw=cyan!80, font=\footnotesize},
    personalization_work/.style =      {align=center, text width=10cm, rounded corners=3pt, line width=0.3mm, fill=cyan!10, draw=cyan!0, font=\footnotesize},
    risk/.style =         {align=center, text width=1.8cm, rounded corners=3pt, line width=0.3mm, fill=orange!10, draw=orange!80, font=\footnotesize},
    risk_work/.style =    {align=center, text width=10cm, rounded corners=3pt, line width=0.3mm, fill=orange!10, draw=orange!0, font=\footnotesize},
}

\usepackage{CJK}

\title{{\fontsize{12.5}{17}\selectfont Building a Scalable, Reproducible, Evaluatable, and Closed-Loop\\
Simulation Environment Foundation for Embodied Intelligence}\\
\normalsize Cloud-Native Simulation Infrastructure for Embodied Intelligence Data Collection, Training and Evaluation 
}

\author{
Junwu Xiong$^{*}$\thanks{Corresponding author. Email: xiongjunwu.1@jd.com},
Yongjian Guo$^{1,2}$,
Mingxi Luo$^{1}$,
Ning Qiao$^{1}$,
Lei Kang$^{1,3}$,
Song Wang$^{1}$,
Yince Gao$^{1,4}$,
Chenfeng Gu$^{4}$,
Zhen Sun$^{4}$,
Haoran Li$^{1}$,
Wei Lu$^{1}$,
Yucheng Guo$^{1}$,
Shuai Di$^{1}$,
Xiaodong Bai$^{1}$,
Haoran Sun$^{1,3}$,
Jing Long$^{1}$,
Jiaxuan Gao$^{1,2}$,
Hui Zhang$^{1}$,
Peng Hao$^{1}$,
Lu lu$^{1}$
}

\affiliation[1]{AI Infra Team at JDT\quad $^2$Tsinghua University\quad $^3$Peking University
\quad $^4$Beihang University}

\abstract{
This white paper presents a cloud-native simulation infrastructure framework for embodied intelligence evaluation and simulation-based data collection. Rather than focusing on a single simulator, dataset, or benchmark, the proposed framework integrates simulation environment generation, task execution, trajectory collection, model evaluation, data governance, closed-loop optimization, and cloud service delivery into a scalable, reproducible, evaluable, and continuously evolving infrastructure foundation.
Driven by the rapid development of vision-language-action models, world models, reinforcement-learning-enhanced VLA methods, and general-purpose robotic systems, embodied intelligence increasingly depends on large-scale interaction data, standardized evaluation, and long-tail failure samples. Although real-robot data remains indispensable, it is constrained by high cost, limited scalability, weak reproducibility, and safety risks. Cloud-native simulation addresses these challenges through resource pooling, elastic scheduling, containerized runtimes, unified data formats, and service-oriented interfaces, enabling simulation as a shared platform for large-scale collaboration.
The white paper proposes a four-layer architecture consisting of the resource and scheduling layer, simulation runtime layer, inference and task execution layer, and data and evaluation service layer. Based on this architecture, it further defines an environment asset and task generation system, a simulation trajectory pipeline, a benchmark system for model evaluation, a data-model-environment closed-loop mechanism, and a cloud-service technology system exposed through APIs, SDKs, and Web consoles. It also discusses the roles of D-VLA, RL-VLA³, Sword, and Pre-VLA in high-concurrency simulation, dynamic scheduling, visual augmentation, real-time data filtering, and closed-loop optimization.
Finally, the paper outlines a phased roadmap from prototype validation and platformization to scaled operation, ecosystem expansion, and real-world validation. It also analyzes key risks, including simulation realism, data quality, evaluation standardization, cloud cost, system complexity, and the sim-to-real gap, together with corresponding mitigation strategies. Overall, cloud-native embodied simulation infrastructure is positioned as a critical foundation connecting simulation environments, model training, data assets, standardized evaluation, and real-robot validation.

}

\date{\today}
\correspondence{Junwu Xiong}

\begin{document}

\maketitle

\section{Background and Motivation}

\subsection{The Development Trend of Embodied Intelligence}

Embodied intelligence is becoming a major direction through which artificial intelligence moves from digital information processing toward interaction with the physical world. Unlike traditional AI systems that mainly process text, images, speech, and other digital content, embodied intelligence systems must perceive objects in three-dimensional environments, understand tasks, reason over relationships, and execute physically constrained actions through robotic bodies. This means that models must not only understand multiple modalities, but also generate executable, verifiable, temporally continuous, and environment-adaptive behavior policies. As AI capabilities extend from cognitive understanding to action execution, embodied intelligence is becoming an important technical path connecting large-model capabilities with real-world applications.

In recent years, the rapid development of vision-language-action models, world models, and reinforcement-learning-enhanced VLA methods has changed the research paradigm of embodied intelligence \cite{brohan2023rt1,brohan2023rt2,kim2024openvla,octo2024,ha2018worldmodels,hafner2020dreamer}. VLA models attempt to unify visual perception, language understanding, and action generation in one modeling framework, enabling robots to understand task goals from natural language instructions and generate actions based on environmental observations. World models focus on modeling environmental dynamics, object interaction relationships, and future states, providing a foundation for task planning, long-horizon reasoning, and self-correction. Reinforcement-learning-enhanced VLA methods further introduce environmental feedback into model optimization, allowing models not only to imitate existing trajectories, but also to improve policies through trial, error, and reward feedback \cite{sun2026rlvla3,guo2026dvla}. At the same time, rapid progress in humanoid robots and general-purpose robotic systems is moving embodied intelligence from laboratory algorithm validation toward real application scenarios.

Under this trend, the development of embodied intelligence no longer depends only on improvements in model architectures or algorithms. It increasingly depends on systematic support from data, environments, and evaluation. Models must be trained and validated across many tasks, scenarios, objects, initial states, and interaction conditions before they can develop stable generalization and reliable execution capabilities. This is especially true for long-horizon tasks, complex manipulation tasks, and open-environment tasks, where progress depends on controllable, reproducible, scalable interaction environments and on systematic collection and assessment of successful, failed, abnormal, and long-tail cases.

The role of simulation environments in embodied intelligence R\&D is therefore changing. Early simulation was often treated as a tool for algorithm debugging and functional validation, used to reduce experiment cost in local tasks \cite{shannon1998simulation}. As embodied models become larger and application scenarios become more complex, simulation is evolving from a standalone tool into infrastructure for model training, data generation, capability evaluation, safety validation, and closed-loop optimization. A simulation system for embodied intelligence must provide not only physical environments and robot execution interfaces, but also batch task execution, automatic trajectory collection, unified model access, metric computation, experiment traceability, and data asset management. In this sense, simulation infrastructure is becoming a key foundation for the continuous iteration and engineering deployment of embodied intelligence models.

From a cloud perspective, building embodied simulation infrastructure has clear strategic value. Simulation tasks for embodied intelligence usually require high compute, high concurrency, and complex data movement, which makes them naturally suitable for cloud-native hosting and scheduling. A cloud platform can integrate distributed simulation environments, compute resources, data storage, task execution, and service interfaces into unified infrastructure, providing algorithm and robotics teams with a stable, scalable, and reusable R\&D foundation. This capability improves the efficiency of individual models and projects while also helping an organization accumulate long-term embodied intelligence capabilities.

\subsection{Bottlenecks in Real Robot Data and Evaluation}

Real robot data is irreplaceable for the development of embodied intelligence models. Lighting changes, object materials, spatial layouts, sensor noise, execution errors, and human-robot interaction patterns in real scenes provide the data distribution closest to deployment. A model can ultimately prove its usability, safety, and robustness only in the physical world. However, as task complexity and model scale grow, relying entirely on real robots for data collection and model evaluation can no longer meet the needs of high-frequency iteration, large-scale validation, and systematic assessment.

First, real robot data collection is costly and slow. Each collection process usually requires coordination among robot hardware, sensors, end effectors, experiment sites, task materials, and human operators. For long-horizon or complex manipulation tasks, each valid trajectory may require environment setup, device debugging, task execution, failure recovery, data organization, and quality inspection. Even for the same task, different operators, device states, and environmental details can lead to different collection results. This makes real robot data difficult to acquire at low cost and at internet-data scale.

Second, real robot experiments have limited execution efficiency and reproducibility. Embodied models must often be tested under many initial states, object layouts, and task constraints, while real experiments are affected by placement errors, manipulator control deviations, sensor noise, lighting changes, and human intervention. If every model update requires large-scale real experiments, R\&D cycles become much longer and automated evaluation becomes difficult. The lack of reproducible environments further weakens the reliability of comparisons across model versions.

Third, real scenes are inefficient at covering long-tail failures. Problems in deployment often appear not in routine successful scenarios, but in occlusion, collision, grasp failure, object slipping, misrecognition, path blockage, task interruption, and environmental disturbance. These cases occur infrequently but are critical for robustness, safety, and generalization. Constructing large numbers of long-tail scenarios on real robots is inefficient and may cause hardware wear or safety risks. Without controllable generation and replay of long-tail failure cases, potential problems are hard to expose before deployment.

Therefore, although real robot data and real-world evaluation are the final basis for embodied intelligence deployment, they cannot alone support large-scale model training, systematic evaluation, and rapid iteration. A more reasonable technical path is to combine real robot experiments with high-fidelity simulation and sim-to-real techniques such as domain randomization and dynamics randomization \cite{tobin2017domain,sadeghi2017cad2rl,peng2018simtoreal}. Simulation systems can undertake large-scale task execution, scene expansion, trajectory collection, failure replay, and standardized evaluation, while real data calibrates the simulation environment and model performance. Cloud-native simulation infrastructure is built to provide a more efficient, controllable, and scalable R\&D and evaluation foundation outside real robot experiments.

\subsection{Value of Cloud-Native Simulation Infrastructure}

The core value of cloud-native simulation infrastructure is to unify dispersed capabilities in simulation execution, task running, data collection, model evaluation, and result management into schedulable, scalable, reproducible, and service-oriented platform capabilities. It is not simply deploying simulation software in the cloud, nor is it a temporary environment for one model or one task. It is a long-term infrastructure for embodied intelligence R\&D that supports multi-model, multi-task, and multi-team collaboration.

From the compute perspective, embodied simulation tasks naturally fit cloud-native architecture. Different scenes, tasks, random seeds, and model versions can often run in parallel, making the workload highly batch-oriented and concurrent. Through resource pooling and elastic scheduling, a cloud platform can allocate GPU, CPU, storage, and network resources on demand to different simulation instances, supporting large-scale parallel simulation, batch trajectory collection, and automated model evaluation. Compared with local workstations or single-machine simulation environments, a cloud-native approach better handles peak workloads, resource isolation, and multi-team sharing, improving overall resource utilization and R\&D throughput.

From the data management perspective, embodied simulation data is multimodal, strongly temporal, and highly relational. A complete simulation task may generate RGB images, depth maps, semantic labels, robot states, joint information, end-effector poses, object poses, action sequences, task descriptions, scene configurations, success/failure labels, logs, and evaluation metrics. If these data are stored across different machines or directories, they become difficult to search, reuse, and trace. Cloud-native infrastructure can provide unified data storage, metadata management, version control, quality inspection, and permission control, allowing trajectory data, task data, and evaluation results to become organization-level data assets.

From the model evaluation perspective, cloud-native simulation infrastructure can provide consistent execution environments and evaluation standards for different models. VLA models and reinforcement learning policies can both access simulation environments through unified interfaces and be tested under the same tasks, scenes, and success criteria. During execution, the platform can automatically record task completion rate, execution steps, collisions, trajectory smoothness, action stability, failure types, resource consumption, and runtime latency, forming comparable, traceable, and reproducible evaluation results. This is valuable for model selection, version iteration, issue localization, and pre-release validation.

From the service access perspective, cloud teams have a natural advantage in building unified platforms and interfaces. Through API, SDK, and Web services, simulation infrastructure can provide standardized access for different R\&D teams. Users can submit collection jobs, invoke scene assets, connect model services, view evaluation results, and download data products through a unified entry point. This lowers the barrier to using simulation systems and reduces duplicated construction and fragmented environments inside the organization. For embodied intelligence, which spans algorithms, robotics, data, platforms, and business, unified cloud infrastructure improves collaboration efficiency and engineering consistency.

From the closed-loop optimization perspective, cloud-native simulation infrastructure can connect model training, simulation evaluation, failure analysis, and data regeneration. When a model fails in certain tasks or scenes, the platform can preserve full trajectories, environment states, model outputs, and failure logs, and then generate similar, perturbed, or more difficult long-tail samples for subsequent training and evaluation. As tasks, data, and evaluation results accumulate, the platform can gradually form a loop of model execution, result evaluation, failure attribution, data supplementation, model optimization, and revalidation. This mechanism helps models learn continuously from failures and improve robustness and generalization in complex environments.

\subsection{Goals and Scope}

This white paper proposes a cloud-native simulation infrastructure framework for embodied intelligence simulation data collection and model evaluation. It does not focus on a single simulator, dataset, or benchmark. Instead, it centers on key links such as simulation environment generation, task execution, trajectory collection, model evaluation, data management, closed-loop optimization, and cloud service delivery, building a scalable, reproducible, evaluable, and closed-loop foundation for embodied simulation environments.

The first goal is to explain why embodied intelligence R\&D needs cloud-native simulation infrastructure. With the development of VLA models, world models, and humanoid robots, embodied models need rapidly growing amounts of interaction data, task evaluation, and failure samples. Traditional approaches based on single-machine simulation or real robot experiments cannot satisfy large-scale parallel execution, standardized evaluation, and high-frequency model iteration. Cloud-native simulation infrastructure can organize compute resources, simulation environments, data collection, task execution, and model evaluation into a stable, scalable, and reusable platform.

The second goal is to propose an infrastructure capability framework for data collection and simulation-based evaluation. The framework should support standardized simulation task definition, batch execution of simulation instances, automatic trajectory collection, unified model service access, automatic metric computation, continuous experiment tracking, and unified data asset management. With these capabilities, simulation environments move beyond algorithm debugging tools and become a platform for model training, model evaluation, data production, and closed-loop optimization.

The third goal is to clarify the current construction stage and future evolution. Based on current project practice, the present stage focuses on engineering adaptation of the data collection link in a new high-fidelity simulation environment, with emphasis on validating task execution, simulation interaction, trajectory recording, and data export. The system can currently support one task running normally in Isaac Sim 5.1 and complete the corresponding trajectory collection. This stage opens the core data link in the new simulation environment and lays the foundation for batch scheduling, multi-task collection, model access and evaluation, data management, and cloud service delivery.

The target readers include cloud platform teams, embodied intelligence algorithm teams, robotics system teams, simulation platform teams, data engineering teams, and technical decision makers. For cloud platform teams, the paper explains the necessity, value, and capability boundaries of embodied simulation infrastructure. For algorithm and robotics teams, it provides a design for an accessible, reproducible, and evaluable experiment and data foundation. At the organizational planning level, it emphasizes accumulating embodied simulation capabilities through cloud-native infrastructure, reducing duplicated construction, improving utilization of data and compute, and supporting continuous evolution across models, tasks, and scenes.

\section{Overall Positioning and Design Principles}

\subsection{Overall Positioning}

For embodied intelligence model evaluation and simulation data collection, the overall positioning of cloud-native simulation infrastructure is to build a cloud simulation foundation integrating environment generation, task execution, trajectory collection, model evaluation, data management, and closed-loop optimization. The platform is not a simple deployment of a simulator, nor a temporary tool for one robot, task, or dataset. It is an infrastructure capability for long-term embodied intelligence R\&D, providing scalable operation, standardized access, reproducible evaluation, and continuous evolution.

Functionally, the platform should cover the full lifecycle of embodied simulation tasks. It should support the generation of simulation environments and task scenes, including robot embodiments, scene assets, object configurations, task goals, initial states, and environmental perturbations. During task execution, it should automatically collect trajectory data such as visual observations, robot states, action sequences, object poses, task logs, and success/failure information. Based on collected data and execution results, it should provide unified metrics, data management mechanisms, and closed-loop optimization processes.

As a platform, the infrastructure should be cloud-based, platformized, and service-oriented. Cloud-based means simulation tasks no longer depend on single machines and manual execution, but use elastic compute, container scheduling, storage management, and resource isolation for large-scale parallel execution. Platformized means simulation environments, task definitions, model interfaces, data formats, and evaluation standards are abstracted into reusable capabilities. Service-oriented means the platform exposes capabilities through API, SDK, or Web services, allowing users to submit tasks, connect models, collect data, and obtain evaluation results in a standard manner.

\subsection{Core Construction Goals}

The core goals of cloud-native simulation infrastructure are to address the main bottlenecks in embodied model R\&D by forming systematic capabilities for large-scale environment generation, standardized model evaluation, high-quality trajectory collection, multi-model/multi-robot/multi-task access, and closed-loop optimization among data, models, and environments.

First, the platform should support large-scale simulation environment generation and execution. Generalization in embodied models depends on diverse environments and task distributions. The platform should not be limited to a few fixed scenes. It should generate diverse environment configurations automatically or semi-automatically around task goals, covering scene layouts, object combinations, initial states, lighting, physical parameters, perturbations, and random seeds. Combined with elastic cloud compute and scheduling, it should support many parallel instances and turn simulation from manual experiments into scaled data production and batch evaluation.

Second, the platform should support high-quality trajectory collection. Embodied trajectory data includes not only visual images or videos, but also robot states, action sequences, joint information, end-effector poses, object states, task semantics, environment configurations, and execution outcomes. High-quality trajectories require temporal completeness, semantic consistency, state traceability, and standardized formats. The platform should automatically record key data during execution and provide cleaning, quality inspection, metadata annotation, and version management.

Third, the platform should support standardized model evaluation. Evaluation should not rely only on one success/failure result. It should include task completion, execution efficiency, trajectory quality, action stability, collision risk, failure types, and resource consumption. The platform should provide consistent task definitions, scene configurations, runtime constraints, and evaluation standards for different models, enabling horizontal comparison of VLA models under unified protocols and tracking version iteration and capability boundaries.

Finally, the platform should create a closed loop among data, models, and environments. Simulation infrastructure should not stop at one-time data collection or one-time evaluation. It should discover failure scenarios, generate supplemental data, adjust task difficulty, and drive retraining and revalidation based on model performance. When a model repeatedly fails in certain tasks, the platform should preserve full trajectories and environment states, then construct similar or perturbed scenes for targeted collection and optimization.

\subsection{Design Principles}

The platform should follow cloud-native, scalable, evaluable, reproducible, and service-oriented design principles. These are not isolated technical requirements; together they support long-term evolution, cross-team reuse, and engineering deployment.

The cloud-native principle requires full use of cloud capabilities such as resource pooling, elastic scheduling, containerized deployment, task orchestration, storage management, and service governance. Embodied simulation tasks have high compute demands and strong parallelism. Through cloud-native architecture, simulation instances, model services, data storage, and evaluation tasks can be included in a unified scheduling system with on-demand allocation, elastic scaling, fault isolation, and runtime monitoring.

The scalability principle requires the platform to support future growth in task scale, model types, robot embodiments, and scene assets. It should not be designed only for the current single-task or single-pipeline collection requirement, but should reserve space for horizontal scale and vertical evolution. Horizontal scale means more parallel tasks, larger data collection, and more teams; vertical evolution means more complex tasks, higher-fidelity physics, richer sensor data, and more complete closed-loop optimization.

The evaluability principle requires metrics and protocols to be built into the core platform from the beginning, rather than added after execution. Embodied model evaluation depends on clear task definitions, success criteria, and metric systems. The platform should automatically compute task completion, efficiency, action quality, safety, stability, and failure causes, and compare results across models, versions, and task conditions.

The reproducibility principle requires complete recording and replay of key conditions for each simulation experiment. Results can be affected by simulator version, scene configuration, random seed, model weights, control frequency, task parameters, and perturbations. The platform should record task configuration, runtime environment, model version, input/output data, and evaluation results through unified experiment metadata management.

The service-oriented principle requires platform capabilities to be exposed through standard interfaces rather than manual operation by platform developers. Algorithm, robotics, and data teams should submit tasks, connect models, query data, download results, and view evaluations through APIs, SDKs, or Web pages. Service orientation lowers usage barriers and turns simulation from a project-local tool into a shared, governable organizational capability.

\subsection{Differences from Traditional Simulation Platforms}

Traditional simulation platforms usually focus on single-machine development, physical simulation, and algorithm debugging. Their core value is providing visual, interactive, and controllable experiment environments for researchers. This is important for early validation, but it exposes limitations when facing large-scale model training, standardized evaluation, and continuous data production: weak scaling, limited automation, scattered data management, and difficult collaboration.

Cloud-native simulation infrastructure differs first in its goal. Traditional platforms ask whether a task can run on a certain machine; cloud-native infrastructure asks whether many tasks can run stably in a shared resource pool, automatically and in parallel, under different models, scenes, and random conditions. The former emphasizes simulator functionality; the latter emphasizes scheduling, scaling, reproducibility, and governance in the cloud.

The second difference is that automated data collection becomes a core capability. Traditional platforms often rely on researchers to organize data manually. Cloud-native platforms record trajectories, states, logs, labels, and metrics during task execution and write them to standardized storage. Such data can support current analysis and become organization-level assets for model training, failure replay, and horizontal evaluation.

The third difference is standardized task protocols and model interfaces. In traditional platforms, task definitions, model calls, and evaluation logic are often coupled in project-specific code. A cloud-native platform abstracts tasks, models, robots, and data formats into unified protocols so that different teams can connect according to consistent specifications. Standard interfaces support multiple models and robot types under the same task system and make evaluation results comparable and traceable.

The fourth difference is platform service delivery. Traditional tools are mainly for local developer use, while a cloud-native platform should provide shared services to multiple internal teams. Users do not need to care how the simulator is installed, how resources are allocated, how data is stored, or how metrics are computed. They submit tasks, view progress, obtain data, and analyze results through a unified entry point.

\section{Industry Status and Technology Trends}

\subsection{Mainstream Embodied Simulation Platforms}

The embodied simulation ecosystem is developing rapidly and forming a multi-layer technology system centered on high-fidelity physics, robot learning environments, task benchmarks, data generation tools, and cloud runtime capabilities. Platforms differ in design goals, physics engines, rendering, robot support, task types, learning interfaces, and engineering ecosystems.

Isaac Sim and Isaac Lab are important ecosystems for high-fidelity robot simulation and robot learning, continuing the trend toward GPU-accelerated robot simulation represented by Isaac Gym \cite{makoviychuk2021isaacgym}. Isaac Sim is based on NVIDIA Omniverse and emphasizes physical realism, sensor simulation, synthetic data generation, robot testing, and validation. It is suitable for visually rich, physically complex, and sensor-heavy robot environments. Isaac Lab builds more standardized robot learning workflows on top of Isaac Sim, including reinforcement learning, imitation learning, motion planning, and task development interfaces. Its advantages are GPU acceleration, high-fidelity rendering, sensor simulation, and industrial ecosystem support. It is suitable for robot foundation models, reinforcement learning policies, sim-to-real transfer, and complex interaction tasks. However, Isaac also has high hardware, environment configuration, and version compatibility requirements, so cloud-scale deployment and standardized multi-team use still require platform-level encapsulation.

MuJoCo is a high-performance physics engine widely used in robot learning and control research \cite{todorov2012mujoco}. It is fast, has stable contact dynamics, offers simple interfaces, and has a mature community. It is widely applied in reinforcement learning, control algorithms, motion planning, and robotic manipulation, especially where high-frequency interaction and large sampling are required. MuJoCo is lightweight and efficient, with strong physical modeling, and has been adopted by many robot learning frameworks and benchmarks. Compared with high-fidelity visual simulation platforms, however, large-scale realistic scene construction, complex visual sensor simulation, and photorealistic rendering are not its core strengths. For VLA models and multimodal embodied data collection, MuJoCo often needs to be combined with other rendering, task wrapping, or data generation tools.

SAPIEN is a physical simulation environment for robotic vision and interaction tasks, emphasizing articulated objects and physical interactions \cite{xiang2020sapien}. With rich interactive object assets and physics simulation, it supports manipulation learning in homes, offices, and daily interaction scenes. Its value lies in combining object parts, joint structures, interactive properties, and robot actions, enabling fine-grained manipulation tasks. For embodied models that must understand object structure, part motion, and interaction constraints, such platforms provide important data and task foundations.

Genesis is a newer platform for physical AI and robot simulation. It emphasizes unified multi-physics engines, high-performance parallel simulation, photorealistic rendering, and scalability for large-scale robot learning. Its emergence reflects an industry shift from competition among traditional simulators toward high-performance, scalable, generative, and service-oriented physical world modeling platforms. Although such platforms are still evolving quickly, their direction indicates that future embodied simulation infrastructure will focus more on GPU parallelism, unified physical modeling, rapid data generation, and deep integration with large-model training pipelines.

Overall, mainstream simulators have built rich ecosystems in physics, rendering, robot control, task construction, and learning interfaces. But most still exist as simulators, algorithm research environments, or benchmark tools, rather than cloud-native infrastructure for organization-level multi-team use. For embodied intelligence evaluation and data collection, the core challenge is not to redevelop a physics engine, but to build unified cloud runtime, task scheduling, data collection, model evaluation, and closed-loop optimization on top of existing simulation ecosystems.

\subsection{Mainstream Benchmarks and Datasets}

As embodied models move from single-task policy learning to general robot capability learning, benchmarks and datasets become increasingly important. Unlike relatively static datasets in vision or language, embodied data is interactive, temporal, and goal-driven. It must record observations, actions, states, language, environments, and outcomes. Current benchmarks and datasets can broadly be divided into robot manipulation, navigation and indoor interaction, VLA tasks, real robot trajectories, simulation trajectories, and multimodal embodied datasets.

Robot manipulation benchmarks are among the most active directions. RLBench, ManiSkill, LIBERO, robosuite, and RoboCasa provide standardized environments for manipulation, long-horizon tasks, multitask learning, lifelong learning, and daily object interaction \cite{james2020rlbench,gu2023maniskill2,liu2024libero,zhu2024robocasa}. RLBench provides many hand-designed vision-guided manipulation tasks and supports expert demonstration generation through motion planners. ManiSkill emphasizes task diversity, object geometry variation, and large-scale demonstration generation. LIBERO focuses on lifelong robot learning and knowledge transfer through language-conditioned tasks and task suites. RoboCasa targets everyday scenes such as home kitchens and emphasizes complex scenes, diverse tasks, and a mixture of human demonstrations and synthetic trajectories.

Navigation and indoor interaction benchmarks target perception, localization, memory, planning, and interaction for mobile agents in complex 3D environments. The Habitat ecosystem supports visual navigation, object-goal navigation, embodied question answering, instruction following, and indoor exploration \cite{kolve2017ai2thor,szot2021habitat}. BEHAVIOR-1K proposes many long-horizon household activities based on real human needs, complex environmental interaction, and multi-step execution \cite{srivastava2022behavior}. Compared with pure navigation, these benchmarks emphasize long-term planning and object interaction in semantic scenes, closer to future service and home robot applications.

VLA and language-conditioned robot tasks have attracted broad attention. Benchmarks such as CALVIN drive long-horizon robot manipulation through language instructions, requiring models to connect language goals, visual observations, and continuous actions \cite{mees2022calvin}. These tasks are closely related to VLA models because they require understanding natural language and generating reasonable behaviors under different environmental states, rather than learning one fixed motion pattern. Language-conditioned tasks are becoming important for evaluating generality, compositional generalization, and long-horizon execution.

Real robot trajectory datasets are important resources for robot foundation models and VLA models. Open X-Embodiment collects large-scale robot data from different robots, tasks, and institutions, proposes a unified data format and RT-X models, and promotes cross-robot and cross-task policy learning \cite{openxembodiment2024}. DROID collects real-world robot manipulation data in a distributed manner, covering diverse scenes, tasks, and operators \cite{khazatsky2024droid}. BridgeData V2 provides many real robot manipulation trajectories and supports language-conditioned, goal-image-conditioned, and multitask learning \cite{walke2023bridgedata}. These datasets show that embodied intelligence is moving from small-scale, single-platform, single-task training toward a cross-platform, cross-scene, cross-task data-driven paradigm.

Simulation trajectory datasets and synthetic data are also growing rapidly. Compared with real robot data, simulation data is low cost, controllable, repeatable, and scalable. It can efficiently generate task trajectories, failure cases, environmental perturbations, and long-tail samples. ManiSkill, RoboCasa, RLBench, CALVIN, and other ecosystems support demonstration generation or trajectory collection to varying degrees. Simulation data can complement missing real data coverage, provide stable and reproducible evaluation conditions, and generate targeted new data based on model failures.

Multimodal embodied datasets are becoming a key direction. Embodied data is no longer just RGB images or action labels. It includes language instructions, visual observations, depth information, robot states, action trajectories, object states, task semantics, environment configurations, and success/failure information. Such data is especially important for training VLA models, world models, and robot foundation models. Future dataset value will depend not only on scale, but also on unified formats, complete metadata, clear task semantics, reliable trajectory quality, and support for reproducible evaluation.

\subsection{Limitations of Current Solutions}

Although the external ecosystem of simulators, benchmarks, and datasets is rich, current solutions still have clear limitations from the perspective of large-scale embodied model R\&D and engineering deployment.

First, simulation platforms remain separated from model training workflows. Many simulators provide good physical environments, rendering, or task interfaces, but training systems, data management systems, and evaluation systems are often built separately by users. For algorithm teams, the simulator is only one part of the data or interaction environment; training, log management, model versioning, data filtering, and result analysis happen elsewhere. Without a unified platform, simulation results do not naturally enter the model training and evaluation loop, making experiments fragmented and hard to trace.

Second, task construction remains costly. Embodied tasks require goals, scenes, objects, robots, initial states, success conditions, failure conditions, control interfaces, and output formats. Even when task APIs exist, building a high-quality, reproducible, scalable task requires simulation, robotics, and engineering expertise. Long-horizon tasks, multi-object interaction, and language-conditioned tasks are even more expensive to construct. High task construction cost limits scene scale, task diversity, model generalization, and evaluation coverage.

Third, inconsistent data formats remain a major obstacle. Platforms and datasets define observations, actions, states, language, camera parameters, robot structures, timestamps, task labels, and success/failure information differently. Even for manipulation trajectories, action spaces, control frequencies, coordinate frames, sensor viewpoints, and metadata structures may vary significantly. Teams training general models or robot foundation models must spend substantial effort on data cleaning, conversion, and alignment.

Evaluation protocols are also inconsistent. Benchmarks have different task definitions, success criteria, sampling methods, and metric systems. Even when two models perform well on similar tasks, differences in environments, initial states, trial counts, random seeds, and failure rules can make direct comparison difficult. Without a unified evaluation protocol, organizations cannot form stable leaderboards, version regression tests, or pre-release validation processes.

Cloud-scale capability is another prominent limitation. Many simulators are designed for local development or single-machine experiments. They may support scripted parallelism, but not native cloud scheduling, resource isolation, elastic scaling, queue management, failure retry, automatic data archiving, and multi-tenant access. When simulation expands from small experiments to large-scale data collection and continuous evaluation, local scripts and manual maintenance cannot guarantee stability or resource utilization.

Existing solutions also lack strong data closed loops. Many platforms can execute tasks and generate results, but lack automated mechanisms from failure cases to new task generation, from evaluation metrics to data supplementation, and from model defects to environment expansion. Failures often require researchers to inspect logs and videos manually before constructing new scenes or data. This cannot support high-frequency iteration and large-scale model optimization.

Finally, current toolchains have limited organization-level collaboration. Different teams may use different simulators, model interfaces, data formats, and experiment scripts, causing duplicated resources and fragmented experience. Even if one team builds a good workflow, it may not become shared infrastructure. Embodied intelligence involves algorithms, robotics, simulation, data, cloud platforms, and business roles; without unified infrastructure, engineering fragmentation becomes severe.

\subsection{Technology Trends}

Embodied simulation is evolving from a single physical simulation tool into systematic infrastructure for large-scale model training, data production, standardized evaluation, and closed-loop optimization. Major trends include automated environment generation, large-scale parallel simulation, world-model-assisted simulation, VLA/RL-VLA closed-loop training, and cloud simulation services.

Automated environment generation will become a key platform capability. Traditional simulation environments rely on manual modeling, manual object placement, and manual task definition. With generative AI, procedural scene generation, 3D asset libraries, and semantic scene editing, future platforms will automatically generate layouts, object combinations, initial states, lighting, and perturbations from task goals. This improves diversity, exposes long-tail issues, and reduces task construction cost.

Large-scale parallel simulation will become foundational for training and evaluation. Embodied models require repeated interaction across many tasks, scenes, and random conditions. Single-machine simulation cannot provide enough throughput. With GPU-parallel physics simulation, cloud scheduling, and containerized runtime, future platforms will emphasize multi-instance concurrency, batch data collection, and automated evaluation.

World-model-assisted simulation is another important trend. Traditional simulation depends on explicit physics engines and manually built rules, while world models learn state transitions, object interactions, and future observations from data \cite{ha2018worldmodels,hafner2019planet,hafner2020dreamer}. Physical simulation and learned world models may become complementary: physics provides controllable, interpretable, reproducible interaction environments, while world models improve long-term prediction, scene completion, counterfactual reasoning, and data generation efficiency.

VLA and RL-VLA closed-loop training will move simulation platforms from evaluation tools to model optimization tools. VLA models unify vision, language, and action; RL-VLA methods introduce environmental feedback \cite{brohan2023rt2,kim2024openvla,sun2026rlvla3,guo2026dvla}. Static datasets alone cannot cover failures in complex environments. Simulation platforms can provide controllable interaction environments and connect execution, failure feedback, reward signals, and trajectory data into a loop, becoming part of continuous model optimization.

Cloud simulation service delivery will also become important. As embodied R\&D scales, the cost of each team maintaining its own simulation stack becomes high. Simulation capabilities will more likely be provided as platform services through APIs, SDKs, or Web interfaces for environment generation, task execution, data collection, model evaluation, and result analysis. This improves resource utilization, ease of use, data standards, task protocols, and governance.

\section{Overall Architecture of Cloud-Native Simulation Infrastructure}

\subsection{Architecture Overview}

Cloud-native simulation infrastructure for embodied intelligence evaluation and data collection is not merely a simulator deployed in the cloud. It is a systematic platform supporting environment generation, simulation execution, model inference, task execution, data collection, and evaluation analysis.

The platform can be divided into four layers: the resource and scheduling layer, the simulation runtime layer, the inference and task execution layer, and the data and evaluation service layer. The resource and scheduling layer provides compute, storage, containers, and task scheduling. The simulation runtime layer provides 3D scenes, robot embodiments, sensors, and physical interaction environments. The inference and task execution layer handles model access, observation construction, action parsing, and closed-loop execution. The data and evaluation service layer handles trajectory collection, data governance, metric computation, failure analysis, and service-oriented output.

The architecture's core value is to organize cloud resources, simulation environments, model inference, and data evaluation into one system so that embodied tasks can run at scale in a reproducible and comparable manner.

\subsection{Resource and Scheduling Layer}

The resource and scheduling layer provides compute resources and runtime guarantees for large-scale simulation. Embodied simulation tasks often require GPU, CPU, storage, and network resources simultaneously. Batch evaluation and data collection require many simulation instances to run in parallel.

The platform should package simulation tasks in containers, including the simulation engine, robot models, scene configurations, task scripts, and dependencies. This ensures task isolation and experiment reproducibility. The platform also needs elastic scheduling to allocate resources based on queues, resource states, and user quotas, enabling batch execution, failure retry, and resource recycling. Multi-tenant isolation, access control, log tracing, and exception recovery are also required so that teams and projects can safely share platform capabilities.

\subsection{Simulation Runtime Layer}

The simulation runtime layer is the virtual world where embodied tasks occur. It provides simulation engines, scene assets, robot embodiments, sensor systems, and physical interaction.

The platform should support multiple engines for different task needs. High-fidelity visual tasks require rendering quality; reinforcement learning tasks require parallel efficiency; complex manipulation tasks depend on contact dynamics and control stability. The platform should hide engine differences behind unified task protocols and data interfaces.

For scenes and robots, the platform should support typical environments such as homes, offices, warehouses, kitchens, retail spaces, hospitals, labs, and industrial production lines, as well as robot forms such as manipulators, mobile robots, dual-arm robots, humanoids, and dexterous hands. Sensor systems should support RGB, depth, multi-view video, point clouds, robot states, and force/tactile information.

\subsection{Inference and Task Execution Layer}

The inference and task execution layer connects models with simulation environments. It should support VLA models, world models, reinforcement learning policies, multimodal agents, and traditional controllers, and should translate model outputs into executable actions in the environment.

This layer includes three main capabilities. Observation construction and input adaptation convert images, depth, states, and task information into model inputs. Action parsing and execution mapping convert language plans, structured actions, skill calls, or continuous control values into robot commands. Closed-loop execution continuously performs the observation, inference, action, and feedback cycle. For long-horizon tasks, the platform should also support subgoal management, execution state tracking, and failure recovery.

\subsection{Data and Evaluation Service Layer}

The data and evaluation service layer turns simulation execution into trainable, analyzable, and comparable data assets. The platform should record task instructions, environment configurations, sensor observations, robot states, model inputs and outputs, action sequences, success/failure labels, and related metadata.

For data governance, the platform should establish unified data formats and quality controls, automatically checking missing images, abnormal actions, object interpenetration, incorrect state labels, and other issues. For model evaluation, it should compute task success rate, subgoal completion rate, action legality, collision rate, trajectory efficiency, instruction following, robustness, and other metrics, then generate standardized reports.

This layer also exposes service capabilities through APIs, SDKs, and Web consoles for task creation, model access, simulation execution, trajectory download, metric viewing, and failure analysis.

\subsection{Chapter Summary}

This chapter proposed a four-layer architecture for cloud-native embodied simulation infrastructure: resource and scheduling, simulation runtime, inference and task execution, and data and evaluation services. It uses cloud resources as the foundation, simulation environments as the carrier, model inference and task execution as the core link, and data accumulation and standardized evaluation as the output.

\section{Simulation Environment Generation and Task System}

\subsection{General Approach}

Training, evaluation, and iteration of embodied models depend on large numbers of high-quality, diverse, interactive environments and tasks. Unlike traditional vision or language tasks, embodied tasks occur in three-dimensional space. Agents must identify objects, understand spatial relationships, determine how objects can be manipulated, and change environment states through actions. A simulation platform for embodied intelligence therefore cannot provide only static 3D scenes; it must have systematic environment generation and task construction capabilities.

The core goal of the environment generation and task system is to organize scene assets, robot embodiments, interactive objects, task templates, and randomization mechanisms into a reusable, composable, and scalable generation system. Starting from finite base assets, the platform can automatically generate many simulation instances with different layouts, object combinations, initial states, and task goals, supporting training, standardized evaluation, failure mining, and closed-loop optimization.

\subsection{Environment Asset System}

The environment asset system is the foundation of task generation. A platform for embodied intelligence should cover common real-world spaces and object combinations, such as homes, kitchens, offices, warehouses, retail stores, hospitals, labs, and industrial lines. Different scenes differ not only visually but also in spatial structure, object functions, interaction modes, and task goals. Assets should therefore form a structured, semantic, and composable library rather than a pile of 3D model files.

Environment assets can be divided into scene assets, robot embodiments, and interactive objects. Scene assets construct the spatial environment where tasks occur. Home and kitchen scenes require counters, cabinets, drawers, sinks, refrigerators, microwaves, ovens, tableware, and food. Office scenes require desks, chairs, monitors, keyboards, folders, printers, and filing cabinets. Warehouses and retail spaces require shelves, bins, products, carts, and pallets.

Robot embodiments define task executors. The platform should support manipulators, mobile robots, dual-arm robots, humanoids, dexterous hands, and different end effectors. The embodiment library should include structural descriptions, joint parameters, kinematics, dynamics, sensor configurations, control interfaces, and capability descriptions such as navigation, grasping, placing, opening doors, pressing buttons, rotating knobs, and carrying objects.

Interactive objects are key assets. Objects are not only visual targets; they are entities that can be manipulated and whose states can change. The platform should uniformly manage graspable objects, openable structures, buttons, knobs, sliders, appliances, tools, and containers. Each object type should include geometry, semantic category, physical properties, interactive parts, and state transition rules. For example, a drawer needs rail constraints and open ranges; a button needs trigger regions and thresholds; a microwave needs door state, button state, operating mode, set time, and display state.

\subsection{Task Generation Mechanism}

The task generation mechanism organizes scenes, robots, and interactive objects into executable tasks. A mature platform should not rely on hand-written task scripts one by one. It should generate tasks in batches through templates and parameters.

Tasks can be divided into atomic tasks, short-horizon tasks, long-horizon tasks, and open-ended tasks. Atomic tasks are basic operation units such as Pick, Place, Move, Open, Close, Press, Rotate, Push, Pull, and Wait. Short-horizon tasks combine a few atomic tasks, such as opening a microwave door, pressing a start button, or picking up a cup and placing it on a table. Long-horizon tasks include multiple subgoals and continuous state changes, such as putting food in a microwave and starting heating, taking a cup from a cabinet and pouring water, or organizing objects on a desk. Open-ended tasks provide only high-level language goals and initial states, requiring the model to plan and execute autonomously.

Task templates are central. Each template should include task goals, preconditions, target objects, action parameters, success conditions, failure conditions, recording methods, and evaluation metrics. A Press task needs button position, pressing direction, and trigger threshold. A Rotate task needs the knob object, direction, target angle, and tolerance. An Open task needs the openable object, target open angle, and completion rule. By changing object types, initial positions, target states, and disturbances, the platform can generate many structurally similar but varied task instances.

In runtime, task generation should be automated, traceable, and reproducible. The flow can be task goal definition, scene selection, object sampling, robot configuration, initial state generation, constraint checking, and task instance publishing. The system first determines goals from natural language, structured templates, or benchmark suites. It then selects scenes, objects, robot embodiments, and initial positions, samples object poses and states, and publishes valid task instances to the queue after constraint and validity checks.

\subsection{Randomization, Perturbation, and Quality Control}

Randomization and perturbation improve generalization and robustness. Real environments contain uncertainty in lighting, object positions, camera viewpoints, and action execution. If a model is trained and evaluated only in fixed, regular, static environments, it may overfit specific scenes and fail to transfer.

Randomization can be visual, spatial, action-level, and state-level. Visual randomization includes lighting intensity, material texture, object color, camera pose, image resolution, and sensor noise. Spatial randomization includes object position, pose, scene layout, robot initial pose, and target regions. Action perturbation includes control errors, execution latency, trajectory deviation, end-effector pose error, and contact uncertainty. State perturbation constructs challenging scenarios such as occluded objects, partially opened doors, untriggered buttons, slipped objects after grasping, or extra distractors.

Perturbation is also useful for generating failure samples. By injecting contact errors, occlusion, target offsets, abnormal states, and execution failures, the platform can generate recovery trajectories and corrective data for robustness training, failure attribution, and closed-loop optimization.

Randomization must not mean unconstrained generation. The platform must ensure physical plausibility and task executability. Target objects cannot be placed outside robot reach; task goals cannot conflict with robot capabilities; scene layouts cannot make tasks inevitably fail. Before task release, the platform should run collision checks, reachability checks, precondition checks, target state checks, and physical plausibility checks. Invalid tasks should be resampled or sent for review.

\subsection{D-VLA: High-Concurrency Distributed Asynchronous Simulation}

To support online evolution of very large embodied AI models, the platform introduces the D-VLA distributed asynchronous framework to address the resource contention between high-fidelity physics simulation and large-parameter model training \cite{guo2026dvla}.

\begin{itemize}
  \item \textbf{Plane decoupling.} D-VLA decouples the simulation data plane from the weight control plane. In high-concurrency environments such as Isaac Sim, visual and physical data generated by simulation are transmitted through a high-frequency data plane, while model weights are distributed through a low-frequency control plane. This reduces interference between the physics engine, such as PhysX, and deep learning frameworks, such as PyTorch/NCCL, in GPU memory bandwidth and communication streams.
  \item \textbf{Four-thread swimlane model.} By building parallel pipelines for sampling, inference, gradient training, and parameter synchronization, D-VLA hides computation and communication. While one instance group performs physics stepping, another performs asynchronous inference with updated weights. Compared with synchronous polling, this architecture can nearly double throughput under a single-node 16-GPU configuration and provides high-concurrency stability for trillion-parameter models.
\end{itemize}

\subsection{RL-VLA\texorpdfstring{\textsuperscript{3}}{3}: Fine-Grained Environment Sharding and Dynamic Scheduling}

To handle uneven compute loads caused by heterogeneous simulation backends, such as MuJoCo-driven libraries and Isaac Sim-driven scenes, the platform integrates the RL-VLA\textsuperscript{3} framework for flexible environment management \cite{sun2026rlvla3}.

\begin{itemize}
  \item \textbf{Simulator-generator decoupling.} RL-VLA\textsuperscript{3} defines three resource groups: simulator, generator, and trainer. Its core advantage is fine-grained environment sharding, which maps a large task batch dynamically to multiple inference nodes. The platform can adjust inference resources according to physical scene complexity, such as the difference between liquid interaction and simple grasping.
  \item \textbf{Asynchronous interaction buffering.} To address unpredictable long-tail latency in physics simulation, such as sudden compute peaks during collision detection, RL-VLA\textsuperscript{3} introduces a dynamic batching scheduler. Simulator observation requests enter a global queue, and the generator processes them flexibly based on batch size or maximum waiting time. This reduces hardware idle time in reinforcement learning and can maintain more than 85\% peak GPU utilization under unstable simulation task sequences.
\end{itemize}

\section{Simulation Trajectory Pipeline for Data Collection}

For embodied models, simulation environments are valuable not only because they provide interactive physical worlds, but also because they continuously produce high-quality trajectory data in a controllable, reproducible, and auditable way. Cloud-native simulation infrastructure should elevate data collection from an experiment byproduct to a core platform capability, so that every task execution becomes a trainable, evaluable, and traceable data asset.

The current construction stage has validated single-task execution and trajectory collection in Isaac Sim 5.1. This chapter abstracts a platform-oriented trajectory pipeline design for future batch collection, multi-model access, and closed-loop data feedback.

\subsection{Data Collection Process}

The basic flow is: task definition, environment initialization, policy execution, trajectory recording, automatic annotation, data quality control, and data ingestion.

The core of the cloud-native trajectory pipeline is to standardize, automate, and service-enable the full process of task definition, environment running, policy execution, data persistence, quality validation, and asset registration. Unlike traditional local simulation that relies on scripts to save results, the platform should determine data product specifications at task submission and then collect, package, and archive data uniformly during execution.

The complete process has seven stages. In the task definition and data contract stage, users submit collection jobs through templates or APIs and declare a data contract in addition to scene, robot, and task parameters. The contract specifies sensor types, sampling frequency, action space, language instruction format, whether to record failures, output paths, and retention policies. The platform generates collection configuration automatically to avoid format drift.

In the environment initialization stage, the platform creates simulation instances, loads scene assets, robot embodiments, and interactive objects, and initializes the physics engine, rendering pipeline, sensors, and control interfaces. Simulator version, scene ID, random seed, and object layout snapshots are written into the trajectory header as metadata for reproducibility.

In the policy execution stage, the policy may come from expert demonstration, rule controllers, pretrained VLA models, RL-VLA online policies, or intermediate demonstrations produced in a Pre-VLA stage. The platform uses a unified observation-action interface to hide differences between simulators and model services. For online inference, it handles model service scheduling, timeout retry, and inference log association.

In the trajectory recording stage, at every control step the platform records multimodal observations, robot states, actions, object states, task progress, and event logs at agreed frequencies. For pipelines such as Sword, it should additionally record contact events, gripper state, and subgoal completion to support imitation learning and failure analysis.

In the automatic annotation stage, the platform generates functional labels from task definitions and logs, including step rewards, subgoal labels, success/failure decisions, collision or out-of-bounds events, phase transition points, and key-frame indexes. For language-conditioned tasks, it associates each step with the corresponding instruction fragment or sub-instruction. For multi-stage tasks, it annotates the current execution phase.

In the quality control stage, the platform runs automatic rule checks and statistical checks before data persistence, marking or filtering physical anomalies, temporal breaks, illegal actions, and incomplete tasks. Quality results are written into metadata and support either high-quality-only ingestion or graded ingestion for later use.

In the ingestion and publishing stage, accepted trajectories and metadata are written to unified storage and registered in the data asset catalog. Users can query by task, scene, robot, policy version, quality level, and failure type through APIs. Training systems can pull versioned data snapshots, and evaluation systems can reference fixed data versions for regression analysis.

\subsection{Data Types}

The pipeline should support RGB/RGB-D, multi-view video, robot joint states, end-effector poses, force/tactile information, action sequences, language instructions, task plans, success/failure labels, and environment metadata.

Embodied trajectory data is multimodal, temporal, and strongly relational. One collection usually covers perception, state, action, semantics, environment, and results. The platform should allow flexible combinations by task and model type while keeping core fields consistent and traceable.

Visual perception data includes RGB images, RGB-D, semantic segmentation, instance segmentation, normal maps, and multi-view video. For VLA and Pre-VLA training, synchronized main and auxiliary views should be supported, with camera intrinsics, extrinsics, distortion parameters, and coordinate frames. High-fidelity environments such as Isaac Sim 5.1 can also provide ground-truth depth, object IDs, and occlusion relationships for supervision and quality checks.

Robot state data includes joint angles, joint velocities, joint torques, end-effector pose, gripper opening state, base pose for mobile robots, and whole-body pose for humanoids. State data should be aligned by unified timestamps and specify coordinate frames and control frequencies.

Action and instruction data can include joint position/velocity/torque commands, end-effector pose deltas, discrete skill calls, or tokenized actions output by VLA models. Language instructions include task-level descriptions, subgoal instructions, scene constraints, and negative constraints. For long-horizon pipelines such as Sword, both high-level task descriptions and stepwise sub-instructions should be recorded.

Object and environment states include poses, velocities, joint states, opening degrees, contact forces, and key physical parameters of interactive objects. Environment metadata includes scene ID, layout version, lighting, random seed, material parameters, and perturbation settings. These fields are important for failure replay, targeted augmentation, and sim-to-real calibration.

Task and process semantics include task ID, task type, subgoal sequence, phase transitions, success criterion triggers, collision events, abnormal exits, and recovery operations. Result and quality labels include task success/failure, subgoal completion rate, execution steps, total duration, collisions, trajectory quality score, quality-pass state, and preliminary failure type. RL-VLA and reinforcement learning should additionally record step rewards, termination signals, and episode returns.

\subsection{Data Format Standard}

The platform should define a unified Observation + State + Action + Instruction + Reward + Metadata format for training, evaluation, and cross-model access. Inconsistent data formats are a major obstacle to scaled use of embodied data. The infrastructure should define a logical schema and support efficient serialized storage so simulation trajectories can directly serve Pre-VLA pretraining, VLA fine-tuning, RL-VLA online learning, and benchmark evaluation \cite{openxembodiment2024}.

\subsubsection{Logical Data Model}

A single trajectory can be abstracted into the following structure.

\begin{itemize}
  \item \textbf{Episode metadata:} trajectory ID, task ID and version, scene version, robot embodiment ID, simulator and version, policy/model ID and version, random seed, start/end time, collection pipeline version, success/failure label, quality status, and associated evaluation run ID.
  \item \textbf{Frame/step data:} each control step contains Observation, State, Action, Instruction, Reward/Done, Events, and Timestamp.
  \item \textbf{Config snapshot:} task YAML/JSON, scene layout, object list, camera configuration, control frequency, success/failure conditions, and perturbation parameters.
  \item \textbf{Derived labels:} subgoal labels, key frames, failure types, and trajectory quality scores generated by automatic annotation modules.
\end{itemize}

This unified paradigm is compatible with mainstream robot learning schemas such as Open X-Embodiment and RLDS-style schemas, reducing adaptation cost for external training frameworks.

\subsubsection{Storage and Exchange Formats}

Dense frame-level data, such as images, point clouds, and high-frequency states, should use columnar or sharded storage such as Parquet, Zarr, WebDataset, MCAP, or custom shards by episode, enabling parallel reads. Episode indexes and metadata should be stored in metadata services or relational/document databases for search by task, scene, quality, time, and model version. Configuration and logs should be stored as JSON/YAML and linked to trajectory paths in object storage.

For external exchange, standard exporters should support LeRobot, RLDS, HDF5, or internal training platform formats. Exports must include format versions and field mapping descriptions to avoid silent mismatch.

\subsubsection{Versioning and Compatibility}

The data schema should use semantic versioning, such as \texttt{schema\_version: 1.2.0}. When fields are upgraded, the platform should provide backward-compatible readers and migration tools. Fixed benchmark evaluation sets should lock schema versions to ensure comparability across collection times.

At the current stage, the platform should first freeze a minimum viable schema, including observation, state, action, instruction, result label, timestamp, and environment metadata, and validate it end to end on the single-task Isaac Sim 5.1 link. Fields such as contact events, subgoals, multi-view data, and failure recovery can be added as Sword and Pre-VLA pipelines expand.

\subsection{Data Quality Control}

Large-scale simulation collection can easily produce large amounts of unusable or low-value data if quality control is missing. The platform should embed quality checks into the pipeline and create a mechanism of quality inspection during collection and traceable records at ingestion.

\subsubsection{Quality Dimensions}

Physical consistency checks include object penetration, abnormal bouncing, joint limit violations, unexplained velocity jumps, and inconsistency between gripper states and contact events. Visual consistency checks include black frames, rendering anomalies, sudden camera parameter changes, depth/RGB misalignment, and frame-rate drops. Temporal integrity checks include missing frames, unordered timestamps, misaligned control and observation steps, and abnormal episode interruption. Action legality checks include joint limit violations, control-frequency violations, NaN/Inf values, and illegal discrete skill calls. Task completion and semantic consistency checks verify true success criteria, consistency between subgoal labels and logs, and matching between language instructions and executed objects. Log and metadata completeness checks verify required metadata, parseable configuration snapshots, and complete model version and seed records.

\subsubsection{Quality Mechanisms}

Automatic rule checks should run before trajectory persistence and mark failed trajectories as \texttt{rejected} or \texttt{debug\_only}. Statistical checks analyze success rate, average length, collision rate, and trajectory length distributions to detect abnormal batches or scene configurations. Comparative checks compare the same task under multiple seeds or policies to find systemic bias. Human spot checks review high-value data such as expert demonstrations, failure recovery, boundary cases, and borderline quality cases, then feed findings back into the rule library. Quality grading can use A/B/C or \texttt{train\_ready}, \texttt{analysis\_only}, and \texttt{rejected}.

\subsubsection{Integration with the Platform}

Quality results should be written into episode metadata and exposed through Web consoles and APIs. For batch collection, the platform should generate reports covering pass rate, failure reason distribution, abnormal scene lists, and recommended fixes. Quality rules should be versioned so each data batch can be traced to the rule set used.

\subsection{Failure Data and Corrective Data}

Embodied data is valuable not only in successful demonstrations, but also in failures, perturbations, recoveries, and corrections. The platform should systematically collect and manage failure data for RL-VLA training, safety constraint learning, error recovery policies, and benchmark regression tests.

Failure data includes execution failures such as grasp failure, excessive collisions, timeout, object slipping, and blocked paths; perception or understanding failures such as wrong target selection, instruction misunderstanding, or tracking loss under occlusion; control failures such as jitter, unsmooth trajectories, joint limit violations, and actuator saturation; and recovery/correction processes that identify an error from a failed state, replan, retry, and eventually succeed or fail again. Such trajectories are especially valuable for closed-loop control and RL-VLA.

The platform should not discard all failed trajectories during quality control. It should preserve them according to the data contract and task type, with structured failure labels. Collection strategies include full recording for evaluation runs and boundary probing, failure oversampling for long-tail cases, dedicated recovery collection from failed snapshots, and controlled adversarial or perturbation injection for generating failures without risking real hardware.

Failure data should connect to the closed-loop mechanism in Chapter 8. Failure trajectories enter error attribution, drive targeted augmentation and new task generation, and form regression subsets for benchmark comparisons. Each failed trajectory should link to failure type, failure step, key frames, model version, and whether it has been used for retraining.

\subsection{Sword: Structure-Guided Style Augmentation and World-Model Simulation}

To address the lack of visual diversity in simulation trajectories, the platform integrates the Sword architecture in the trajectory pipeline, using world models as neural simulators to expand high-value trajectory assets \cite{gao2026sword}.

\begin{itemize}
  \item \textbf{Style-robust augmentation.} Sword uses structure-guided style augmentation to derive many parallel worlds from existing trajectories at low cost. During recording, Sword uses a Cosmos-Transfer-style architecture to change materials, lighting, and scene textures while keeping task geometry, such as depth maps and segmentation masks, unchanged. A single raw trajectory can be augmented into a mixed dataset containing hundreds of out-of-distribution visual variations.
  \item \textbf{Dynamic latent bootstrapping.} To reduce drift in long-horizon trajectory generation, Sword introduces dynamic latent bootstrapping. During world-model generation, the system periodically calibrates differences between predicted distributions and physical references. This enables Sword to generate hundreds of frames with strong physical consistency, providing large-scale visually robust imitation learning samples for VLA models and filling the gap left by high-cost, slow physical simulation sampling.
\end{itemize}

\subsection{Pre-VLA: Preemptive Trajectory Verification and Real-Time Filtering}

To ensure data quality and reduce invalid rendering cost, the platform embeds Pre-VLA real-time verification in the policy execution stage \cite{sun2026prevla}.

\begin{itemize}
  \item \textbf{Preemptive runtime verification.} At each control step, Pre-VLA's ARGUS verification head performs an online health check on candidate actions generated by the VLA model. Based on historical trajectories and current observations, it predicts whether an action sequence may trigger physical failure, such as irreversible kinematic collision or object dropping. If the score falls below a safety threshold, the platform intercepts the command and triggers preemptive resampling before the simulator renders the next frame.
  \item \textbf{Data pipeline efficiency optimization.} Pre-VLA changes the passive collect-then-label mode. It grades data during trajectory generation: only verified high-value and high-safety trajectories enter high-fidelity rendering and final storage. Experiments show that this mechanism can improve average success rates on tasks such as LIBERO by more than 6.8\%, reduce invalid rendering steps by 20\%-30\%, and improve cloud GPU collection efficiency.
\end{itemize}

\section{Benchmark System for Model Evaluation}

The evaluation capability of simulation infrastructure determines whether the platform can support model selection, version iteration, capability boundary analysis, and pre-release validation. A benchmark should not be understood as a static leaderboard, but as continuously evolving evaluation infrastructure consisting of task suites, metrics, protocols, execution engines, and reporting systems.

\subsection{Evaluation Objects}

The cloud-native benchmark system should cover major model forms likely to access simulation, including VLA models, RL-VLA models, world models, robot policy models, multimodal agents, and traditional control policies.

VLA models take visual observations and natural language instructions as input and directly output actions or action tokens. Evaluation focuses on instruction following, multi-step execution, object generalization, and long-horizon stability. RL-VLA and reinforcement-learning-enhanced policies introduce environmental feedback, rewards, and online/offline policy optimization on top of VLA. Evaluation must consider not only task success but also sample efficiency, exploration safety, failure recovery, and train-evaluation consistency.

Pre-VLA or intermediate models, such as hierarchical planners, skill selectors, and subgoal generators, should support both module-level evaluation and end-to-end joint evaluation. World models focus on state prediction, future observation generation, physical consistency, and long-term rollout stability, with tasks such as one-step prediction, multi-step rollout consistency, contact event prediction, and counterfactual scene reasoning. Robot policy and imitation learning models, such as behavior cloning, diffusion policies, and ACT, are usually evaluated on in-distribution performance and out-of-distribution generalization \cite{chi2023diffusionpolicy,zhao2023act}. Multimodal agents and traditional control baselines, including planning-plus-control agents, motion planners, scripted policies, and PID/impedance control, provide lower and upper reference bounds.

Each evaluation object should have a model descriptor recording model type, input modalities, action space, inference latency requirements, history-window dependence, and weight version, enabling the evaluation system to select suitable task subsets and metric sets automatically.

\subsection{Evaluation Task Hierarchy}

To avoid hiding capability differences behind a single success rate, the benchmark should use a hierarchical task system from atomic abilities to open tasks. A five-level L1-L5 structure should align with the task template system in Chapter 5.

\begin{longtable}{@{}p{0.10\linewidth}p{0.18\linewidth}p{0.34\linewidth}p{0.30\linewidth}@{}}
\toprule
\textbf{Level} & \textbf{Name} & \textbf{Example Tasks} & \textbf{Main Capabilities} \\
\midrule
\endhead
L1 & Atomic actions & Reach, Grasp, Place, Push, Rotate, Open/Close & Basic control precision, action legality, single-step perception \\
L2 & Composite operations & Pick-and-place, drawer opening/closing, button operation, optional bimanual coordination & Short sequence execution, contact robustness, subgoal connection \\
L3 & Long-horizon tasks & Multi-step meal preparation, desk organization, multi-room pickup and delivery & Long-term state tracking, planning consistency, error accumulation \\
L4 & Multi-object or multi-constraint tasks & Sorting objects, placing by attributes, obstacle-avoiding manipulation & Spatial reasoning, object relationship understanding, distractor handling \\
L5 & Open or language-conditioned tasks & Natural-language-driven tasks, unseen object combinations, open-ended goals & Generalization, instruction composition, zero/few-shot adaptation \\
\bottomrule
\end{longtable}

Each level should contain a core task set for mandatory fixed-version testing and an extended set for optional capability probing. Tasks should bind standard scenes, robot embodiments, initial-state sampling rules, and success criteria. The platform should support task suites such as \texttt{Manip-Core-v1} and \texttt{Kitchen-LongHorizon-v1}. At the current stage, the platform can start with a small fixed L1-L2 task set in Isaac Sim 5.1 before expanding to long-horizon and language-conditioned tasks.

\subsection{Metric System}

The benchmark metric system should cover task results, execution process, safety, efficiency, and generalization, rather than relying only on success rate.

\subsubsection{Core Result Metrics}

\begin{itemize}
  \item \textbf{Success rate:} success ratio under fixed trial counts and random seed policies.
  \item \textbf{Subgoal completion rate:} stage-level completion in multi-stage tasks.
  \item \textbf{Instruction following accuracy:} compliance with target objects, order, and constraints in language-conditioned tasks.
\end{itemize}

\subsubsection{Process and Quality Metrics}

Process and quality metrics include execution steps or path efficiency relative to expert or optimal baselines, trajectory smoothness, action legality rate, collision rate and severe collision counts, and recovery success rate after failures within bounded steps.

\subsubsection{Robustness and Generalization Metrics}

Robustness and generalization metrics include success rate under domain randomization, performance drop under lighting/texture/layout/noise perturbations, object generalization success rate, sensitivity to initial states, and pass rate on long-tail regression sets derived from historical failures.

\subsubsection{System and Deployment Metrics}

System metrics include single-step inference latency, end-to-end cycle time, GPU/CPU consumption, and resource profiles. Sim-to-real consistency should be recorded separately as real-robot retest success on key tasks and should not be mixed with pure simulation metrics.

\subsubsection{Layered Usage Recommendations}

L1-L2 should emphasize success rate, action legality, collision rate, and efficiency. L3-L5 should add subgoal completion, instruction following, recovery, and robustness. World models should add prediction error, rollout drift, and contact-event recall. RL-VLA should add sample efficiency, training curve convergence, and safety violations. All metrics must have unified definitions, including how timeouts and collisions count, trial counts, and seed policies.

\subsection{Standardized Evaluation Protocol}

Reproducible and comparable evaluation depends on strict protocols. The platform should solidify evaluation protocols as platform capabilities rather than scattered scripts.

Environment and version locking should fix simulator versions, scene asset versions, robot URDF/USD versions, physical parameters, and rendering settings. Each benchmark run should generate an environment bill of materials and write it into the report. Task and sampling protocols should define trial counts, random seed policies, initial-state sampling ranges, early stopping, maximum steps, timeout handling, and retry rules. Core task success criteria should not change arbitrarily within a major version.

Model access protocols should unify Observation APIs and Action APIs, including image resolution, control frequency, action-space normalization, and history length. The evaluation system should record inference mode, such as local container, remote service, or offline batched policy, and store model weight checksums and inference configuration. During evaluation, models should not modify task definitions, access unauthorized ground truth, or change success criteria unless the suite explicitly allows oracle baselines.

Logging and audit must record observations, actions, state transitions, events, resource usage, and decisions for each trial. Each run should have a unique \texttt{eval\_run\_id} traceable to data format version, platform version, and scheduling node. Protocols should be versioned, such as \texttt{eval\_protocol\_v1.0}; model comparisons should occur under the same protocol version, while cross-version comparisons require bridge tasks or explicit incomparability labels.

\subsection{Evaluation Reports and Leaderboards}

Evaluation outputs should become visual evidence for model iteration and engineering decisions. The platform should provide standardized reports, interactive analysis, and internal leaderboards.

A standard report should include model information, protocol version, task suite, overall metrics, hierarchical results, task-level details, confidence intervals, average steps, failure distributions, comparison with previous versions or baselines, typical failure episodes, key frames or videos, inference latency, simulation runtime, resource consumption, environment BOM, seed lists, and full configuration snapshots. Reports should export to PDF, HTML, and JSON, with JSON enabling CI and model management systems.

Leaderboards should be split by task suite and protocol version to avoid mixing different experiment settings. They should display core metrics such as success rate, efficiency, and collision rate, as well as model metadata such as type, parameter count, and whether extra data is used. Both private team boards and reviewed publication boards can be supported, with historical records for checkpoint comparisons. The leaderboard is a tool, not the goal; the platform should emphasize robust regression and verified failure repair over superficial ranking.

Replay and diagnostic tools should reproduce evaluation episodes in a Web console or desktop tool, support stepwise inspection of observations, actions, and events, extract key frames before failure, compare two models under the same seed, and resubmit reproduction tasks from \texttt{eval\_run\_id} and episode ID.

At the current stage, recommended priorities are one core suite with three to five L1-L2 tasks, a minimal metric set of success rate, steps, collisions, timeouts, and illegal actions, a v0.1 protocol using fixed seeds and 10-20 episodes, JSON reports with a simple Web list, and linkage between evaluation runs and trajectories through \texttt{eval\_run\_id}.

\section{Data-Model-Environment Closed-Loop Mechanism}

This chapter answers how the platform continuously improves data, models, and environments during use. For cloud-native embodied simulation infrastructure, value should not stop at one task execution, one data collection, or one model evaluation. It should form a closed-loop iteration mechanism among data, models, and environments. Through this mechanism, the platform discovers failure samples and capability boundaries from model execution, converts failures into new training data and evaluation tasks, and continuously drives the joint evolution of model capability, task coverage, and environment complexity.

The core process is model evaluation, failure analysis, task resampling, data augmentation, model training, and reevaluation. A model executes standardized simulation tasks while the platform records trajectories, environment states, model inputs and outputs, logs, and metrics. The platform classifies successful, failed, abnormal, and boundary samples and attributes failures structurally. Based on attribution, it generates similar tasks, perturbed tasks, recovery tasks, and harder tasks for datasets or benchmark expansion. After retraining or policy optimization, the model is evaluated again to validate improvement.

\subsection{Error Attribution}

Error attribution is key to the closed loop. Embodied model failures are often caused by multiple factors, including perception, language understanding, task planning, action control, physical interaction, and environment generalization. If the platform records only success or failure, it cannot guide data augmentation or model optimization.

The platform should build attribution across multiple dimensions. Perception errors include incorrect recognition of objects, positions, poses, spatial relations, or states, often caused by occlusion, reflective materials, or sensor noise. Language understanding errors include misunderstanding task instructions, constraints, target objects, or action order. Planning errors occur when the model understands the goal but fails to generate reasonable intermediate steps or subgoals, especially in long-horizon, multi-object, and open-ended tasks. Control errors include unstable, discontinuous, illegal, or non-executable actions, such as joint limit violations, abnormal end-effector poses, excessive motion, or control-frequency mismatch. Contact and physical interaction errors include bad contact point selection, wrong gripper timing, excessive collision, object slipping, or insufficient force. Environment generalization errors occur when models work in familiar scenes but fail under new objects, layouts, lighting, materials, or initial states.

The platform should use logs, state changes, action sequences, collision events, object pose changes, task progress, and key frames for automatic or semi-automatic attribution. Early systems can combine rules and human labels; later systems can introduce model-assisted clustering, label prediction, and similar-case retrieval.

\subsection{Targeted Data Augmentation}

Targeted augmentation converts failure samples into model improvement resources. Unlike generic random augmentation, it should be generated around real model weaknesses, long-tail scenarios, and boundary states.

For perception errors, the platform can generate viewpoint perturbations, lighting perturbations, occlusions, material changes, background changes, and camera noise. For language errors, it can generate alternative expressions for the same task, including attribute changes, spatial relationship changes, target substitutions, ordering constraints, and negative constraints. For planning errors, it can generate multi-stage tasks, explicit-subgoal tasks, long-dependency tasks, and optional-path tasks, while recording decomposition and intermediate states. For control errors, it can collect smoother demonstrations with clearer control boundaries and more stable frequencies, while retaining illegal actions, failed actions, and recovery actions. For contact failures, it can generate trajectories with varied object shapes, contact positions, friction parameters, and grasp poses, covering grasp failure, slip recovery, realignment, and retry.

Failure data itself is valuable. The platform should systematically preserve failed trajectories, corrective trajectories, and recovery trajectories, not only successful ones. For RL-VLA, reinforcement learning policies, and closed-loop control models, failures can support reward modeling, policy improvement, error recovery, and safety constraint learning.

\subsection{Training Environment Integration}

To support closed-loop optimization, simulation environments and model training systems should be decoupled but connected. The simulation system provides controllable interaction environments, trajectory data, and evaluation feedback. The training system handles parameter updates, policy optimization, and model version management. They should connect through standardized data, task, and model interfaces.

For offline training, the platform exports trajectories in unified formats for VLA models, robot foundation models, world models, and imitation learning policies. Exported data should include observations, states, actions, language instructions, task labels, success/failure information, and environment metadata. For online interaction, the platform provides simulation sampling environments for reinforcement learning and RL-VLA. Models execute actions, and the platform returns observations, states, rewards, and termination signals, supporting parallel sampling, failure retry, and environment reset. For automatic evaluation triggers, a newly trained model can call fixed suite evaluation, after which reports are written back to model management systems.

\subsection{Continuous Iteration}

The data-model-environment closed loop should form a technical flywheel. As more models access the platform, the platform accumulates more behaviors and failures. As failure samples increase, the platform generates higher-value data and more discriminative tasks. As data and task systems improve, training and evaluation improve. As models become stronger, the platform exposes harder, longer-horizon, and more open task problems.

This flywheel is especially important for embodied intelligence because open physical worlds have much higher task complexity, environmental variation, and failure diversity than static vision or language tasks. A single dataset or fixed benchmark cannot cover model capability boundaries for long. Only continuous execution, collection, evaluation, and feedback can keep task and data systems aligned with model progress.

\section{Cloud-Service Simulation Technology System}

This chapter answers how simulation capabilities can become reusable, accessible, and scalable technical services in a cloud-native manner. Embodied intelligence R\&D involves simulation engines, robot embodiments, scene assets, physical interaction, model inference, task scheduling, trajectory collection, data storage, and evaluation analysis. If each team maintains these capabilities independently, the result is duplicated environments, inconsistent interfaces, low resource utilization, weak data accumulation, and poor reproducibility. Cloud-native embodied simulation infrastructure should therefore form a service-oriented technology system that encapsulates complex low-level capabilities into unified interfaces.

Service orientation here does not mean presenting the platform as a commercial product. It means abstracting simulation capabilities into callable, orchestratable, and governable infrastructure capabilities. Similar to the API-oriented infrastructure idea represented by Tinker, the platform should hide runtime environments, resource scheduling, and distributed engineering complexity as much as possible. Users describe task goals, model interfaces, and data requirements through APIs, SDKs, or consoles, and the platform automatically creates simulation instances, allocates resources, executes tasks, archives data, and generates reports.

\subsection{Overall Architecture of Service-Oriented Simulation}

The cloud-service simulation technology system can be divided into the interface layer, service orchestration layer, platform capability layer, simulation runtime layer, and resource foundation layer.

The interface layer faces users and upper systems, providing APIs, SDKs, command-line tools, and Web consoles. Users create environments, submit tasks, connect models, query data, and view evaluations through this layer. It should remain stable and prevent users from directly depending on simulator details.

The service orchestration layer manages task lifecycles, including parsing, validation, queue scheduling, instance creation, monitoring, failure retry, state synchronization, and result callback. For batch collection and large-scale evaluation, it should support asynchronous execution, task sharding, concurrency control, and priority scheduling.

The platform capability layer provides common modules such as the task system, data pipeline, evaluation system, model adaptation, experiment management, permission management, and log tracing. This layer is the key difference from a single simulator. It turns common engineering capabilities in simulation execution into reusable platform modules.

The simulation runtime layer hosts concrete engines, scene assets, robot embodiments, sensors, physical interaction, and control interfaces. The platform should support ecosystems such as Isaac Sim, MuJoCo, SAPIEN, Genesis, Habitat, and Gazebo/ROS under a unified runtime framework \cite{makoviychuk2021isaacgym,todorov2012mujoco,xiang2020sapien,szot2021habitat}. Different engines can be used for different task types, while upper layers maintain unified task and data protocols.

The resource foundation layer provides GPU, CPU, storage, network, containers, images, schedulers, and monitoring. It must support resource pooling, elastic scaling, task isolation, multi-tenant management, and cost statistics.

\subsection{Cost, Efficiency, and Stability Optimization}

The service-oriented simulation system must balance cost, efficiency, and stability. Embodied simulation includes resource-intensive rendering and physics, as well as high-throughput data collection and evaluation, so systematic optimization is required.

For compute efficiency, the platform should classify resources by task type. Lightweight control tasks can run on low-cost resources; high-fidelity visual tasks can use GPU rendering instances; large-scale reinforcement learning sampling can use parallel simulation instances. Image prewarming, environment caching, instance reuse, and task sharding can reduce startup and repeated initialization costs.

For data efficiency, the platform should use hot/cold tiered storage. Raw videos, multi-sensor data, and complete logs can stay in hot storage for a limited period, while high-value trajectories, failure samples, evaluation results, and metadata move to long-term storage. Low-value intermediate files can be compressed, archived, or cleaned periodically. Data APIs should support on-demand download and field-level filtering to avoid repeated full transfers.

For stability, the platform should provide health checks, failure retry, checkpoint recovery, log aggregation, alerts, and version rollback. Simulator crashes, model service timeouts, resource shortage, data write failures, and network errors must be handled in standardized ways. Benchmark and regression tests should use fixed environments and version locking to ensure reproducibility.

\section{Technology Roadmap, Risks, and Outlook}

This chapter answers how cloud-native embodied simulation infrastructure can be implemented and how it should evolve. Platform construction should follow a path from single-link validation to platformization, from scaled execution to ecosystem expansion. Because embodied simulation involves high-fidelity physics, multimodal data collection, robot control, model evaluation, cloud scheduling, and real-world transfer, the roadmap should emphasize phased validation, layered decoupling, continuous iteration, and risk closure.

\subsection{Phase 1: Prototype Validation}

The goal of Phase 1 is to run the minimum viable link and validate task execution, trajectory collection, and basic evaluation in the new simulation environment. This phase does not pursue scale or system completeness; it proves that key links can run stably.

The platform should select one main simulation engine, one representative robot embodiment, and a few high-value task scenes. It should complete environment loading, robot initialization, task startup, policy execution, state feedback, trajectory recording, data export, and basic metric computation. For the current construction, priority can be given to validating a single manipulation task in a specified high-fidelity simulation version and saving complete trajectories.

Minimum data format and minimum metrics should be defined. The data format should include observations, states, actions, task instructions, reward or result labels, timestamps, and environment metadata. Metrics should include task success, execution steps, runtime, abnormal exit, collision events, and logs. Key outputs include runnable simulation examples, basic trajectory collection, initial schema definition, basic evaluation scripts, and initial reports. Acceptance requires stable single-task execution, automatic data saving, automatic result decisions, and reproducible configurations.

\subsection{Phase 2: Platformization}

Phase 2 abstracts prototype links into platform capabilities, establishing task systems, data pipelines, model access protocols, metrics, and visual management.

The task system should separate task definitions from individual scripts and form standard templates supporting scene, robot, object, language instruction, initial state, random seed, success/failure condition, and output requirements. The data pipeline should establish automatic collection, quality checks, metadata management, data ingestion, and data export. It should preserve success, failure, abnormal, and corrective trajectories and support search by task, model, scene, time, quality, and failure type.

For model access, the platform should define a unified observation-action protocol and support online inference services, containerized models, and offline policies. Evaluation should build basic metrics around success rate, subgoal completion, execution efficiency, collision rate, action legality, failure type, and resource consumption. Visualization should provide task submission, monitoring, logs, trajectory replay, data search, and reports. Acceptance requires multi-task execution, multi-model access, automated trajectory collection, basic quality checks, and standard reports.

\subsection{Phase 3: Scaled Operation}

Phase 3 supports large-scale parallel simulation, multi-user access, standardized benchmarks, automated quality checks, and model version regression tests. The platform upgrades from a usable system to organization-level infrastructure.

For resource scheduling, the platform should support GPU/CPU pooling, multi-task queues, elastic scaling, task priorities, failure retry, resource quotas, and multi-tenant isolation. Tasks should have resource profiles based on compute load, rendering requirements, data output, and runtime, and use differentiated scheduling strategies.

For benchmarks, the platform should build fixed task sets covering atomic actions, composite operations, long-horizon tasks, multi-object tasks, and open tasks. Each task set should have unified task definitions, seed strategies, input/output formats, success criteria, and report templates. For data production, the platform should support batch trajectory collection, automatic quality checks, failure mining, long-tail scene generation, and data versioning. It should move from data quantity toward data value density, prioritizing failed, recovered, boundary, and difficult task samples.

For model evaluation, the platform should support leaderboards, version regression tests, and failure statistics. After each model version is trained, fixed-suite evaluation can run automatically and compare against historical versions. Acceptance requires large-scale concurrent tasks, multi-user/project isolation, automated benchmark execution, model version comparison, and automatic failure data feedback.

\subsection{Phase 4: Ecosystem Expansion and Real-World Validation}

Phase 4 expands simulation ecosystems, robot embodiments, task scenes, and real-world validation links, turning the platform into a general technical foundation for multiple scenarios, models, and robot forms.

For simulation ecosystems, the platform should gradually support multiple engines and task frameworks. Different simulators have different advantages in rendering, physical interaction, reinforcement learning sampling, navigation, manipulation, and multi-agent tasks. An adaptation layer should hide low-level differences while preserving unified task, data, and evaluation interfaces.

For robot embodiments, the platform should expand from single manipulators to dual-arm robots, mobile robots, mobile manipulators, dexterous hands, and humanoids. Different embodiments have different action spaces, control interfaces, sensor configurations, and task constraints, so embodiment abstraction and control adaptation layers are required.

For real-world validation, the platform should map simulation tasks to real robot tasks. Key tasks, typical failure samples, and pre-release evaluation tasks should enter real robot validation. Real execution results should calibrate simulation parameters, correct task decisions, and optimize data distributions to gradually narrow the sim-to-real gap.

\subsection{Risks and Countermeasures}

Simulation realism is the first risk. High-fidelity rendering does not guarantee accurate physical interaction. Contact, friction, materials, sensor noise, and execution error may differ from the real world. Countermeasures include introducing real robot data for simulation parameter calibration, using domain randomization and perturbation injection to improve robustness, and separating simulation results from real validation results in reports \cite{tobin2017domain,peng2018simtoreal}.

Data quality is the second risk. Large-scale simulation can generate many low-quality trajectories, including abnormal exits, physical inconsistency, annotation errors, illegal actions, and incomplete tasks. Countermeasures include automatic quality checks across physical consistency, visual consistency, action legality, task completion, and log anomalies, combined with human spot checks for high-value data.

Evaluation standardization is the third risk. If teams use different task definitions, random seeds, success criteria, and metrics, results are hard to compare. Countermeasures include unified benchmark task sets, input/output protocols, report templates, and version locking.

Cloud cost is the fourth risk. High-fidelity simulation and large-scale collection consume large amounts of GPU, storage, and network resources. Countermeasures include resource-tiered scheduling, queue management, on-demand startup, idle release, image prewarming, instance reuse, data compression, and hot/cold storage. Cost statistics and quotas should also be established.

System complexity is the fifth risk. Simulation infrastructure involves simulators, containers, scheduling, model services, data systems, evaluation systems, and visualization. Countermeasures include modular architecture, standard interfaces, health checks, failure retry, log tracing, alerts, and version management.

The sim-to-real gap is a long-term challenge. Good simulation performance does not necessarily imply real-world usability. The platform should position simulation as a tool for large-scale training, evaluation, and risk exposure, not as the only deployment evidence. Real robot execution data should continuously calibrate the simulation environment and evaluation metrics.

\subsection{Future Outlook}

Cloud-native embodied simulation infrastructure will continue to evolve toward automated environment generation, world-model-assisted simulation, VLA/RL-VLA closed-loop training, multi-robot collaboration, humanoid robot benchmarks, and standardized cloud simulation services \cite{gao2026sword,sun2026rlvla3,guo2026dvla}.

Automated environment generation will lower task construction cost by combining procedural generation, 3D asset libraries, and generative AI to create layouts, object combinations, lighting, perturbations, and long-tail samples from task goals. World-model-assisted simulation will make simulation systems more intelligent by complementing physical simulation with learned state prediction, counterfactual reasoning, failure risk estimation, and scene completion.

VLA and RL-VLA closed-loop training will push simulation platforms from evaluation tools toward model optimization infrastructure. Models execute tasks in simulation, the platform records failures and generates supplemental data, and models improve through offline training or online reinforcement learning. As this process matures, simulation platforms will become deeply embedded in training and capability iteration.

Multi-robot collaboration and humanoid benchmarks will expand task boundaries. Future tasks will move beyond single-arm manipulation to mobile manipulation, multi-robot cooperation, bimanual coordination, dexterous manipulation, and humanoid long-horizon tasks. The platform must support more complex embodiment abstractions, task protocols, and safety constraints.

Overall, building cloud-native embodied simulation infrastructure is a long-term evolution process. The short-term goal is to open the data collection and basic evaluation link in high-fidelity simulation environments. The mid-term goal is to form platformized, scaled, and standardized capabilities. The long-term goal is to build embodied intelligence infrastructure that connects simulation environments, model training, data assets, standard evaluation, and real-world validation. Through this roadmap, the platform can support embodied intelligence models as they move from algorithm validation to systematic evaluation, data closed loops, and engineering deployment.



\nocite{koenig2004gazebo,quigley2009ros,brockman2016openai,tassa2018dmcontrol}
\nocite{zhu2020robosuite,coumans2016pybullet,yu2020metaworld,xia2018gibson}
\nocite{shen2021igibson,chang2017matterport3d,straub2019replica,deitke2022procthor}
\nocite{savva2019habitat,anderson2018vln,shridhar2020alfred,padmakumar2022teach}
\nocite{mahler2017dexnet,levine2016endtoend,kalashnikov2018qtopt,dasari2019robonet}
\nocite{mandlekar2021robomimic,hafner2023dreamerv3,schrittwieser2020muzero,reed2022gato}
\nocite{ahn2022saycan,driess2023palme,huang2022inner,liang2023codeaspolicies}

\bibliographystyle{unsrt}
\bibliography{main}

\clearpage



\end{document}